\documentclass[journal]{IEEEtran}

\usepackage{graphicx}
\usepackage{amssymb,amsmath,epsfig}
\usepackage{algorithm}
\usepackage{algorithmic}
\usepackage{booktabs}
\usepackage{mathrsfs}
\usepackage{color}
\usepackage{textcomp}
\usepackage{bbding}
\usepackage{pifont}
\usepackage{wasysym}
\usepackage{amssymb}
\usepackage{subcaption}
\usepackage{cite}
\usepackage{amsmath,amssymb,amsfonts}
\usepackage{xcolor}
\usepackage{multirow}
\usepackage{epsfig}
\usepackage{flushend}
\usepackage[hyphens]{url}

\begin{document}
%
\title{FBSNet: A Fast Bilateral Symmetrical Network for Real-Time Semantic Segmentation}
%
%
%
\author{Guangwei Gao\textsuperscript{$\dagger$},~\IEEEmembership{Member,~IEEE,}
        Guoan Xu\textsuperscript{$\dagger$},
        Juncheng Li\textsuperscript{$\ast$},
        Yi Yu\textsuperscript{$\ast$},~\IEEEmembership{Senior Member,~IEEE,}\\
        Huimin Lu,~\IEEEmembership{Senior Member,~IEEE}
        and Jian Yang,~\IEEEmembership{Member,~IEEE}
\thanks{\textsuperscript{$\ast$}Corresponding author, \textsuperscript{$\dagger$}Equal contribution.}
\thanks{This work was supported in part by the National Natural Science Foundation of China under Grant nos. 61972212, 61772568, and 61833011, the Natural Science Foundation of Jiangsu Province under Grant no. BK20190089, the Six Talent Peaks Project in Jiangsu Province under Grant no. RJFW-011.}
\thanks{G. Gao and G. Xu are with the Institute of Advanced Technology, Nanjing University of Posts and Telecommunications, Nanjing, China, and also with the Digital Content and Media Sciences Research Division, National Institute of Informatics, Tokyo, Japan (e-mail: csggao@gmail.com, xga\_njupt@163.com).}
\thanks{J. Li is with the Center for Mathematical Artificial Intelligence, Department of Mathematics, The Chinese University of Hong Kong, Hong Kong, China (e-mail: cvjunchengli@gmail.com).}
\thanks{Y. Yu is with the Digital Content and Media Sciences Research Division, National Institute of Informatics, Tokyo, Japan (e-mail: yiyu@nii.ac.jp).}
\thanks{H. Lu is with the Department of Mechanical and Control Engineering, Kyushu Institute of Technology, Kitakyushu, Japan (e-mail: dr.huimin.lu@ieee.org).}
\thanks{J. Yang is with the School of Computer Science and Technology, Nanjing University of Science and Technology, Nanjing, China (e-mail: csjyang@njust.edu.cn).}
}

\markboth{IEEE Transactions on Multimedia}%
{Shell \MakeLowercase{\textit{et al.}}: Bare Demo of IEEEtran.cls for IEEE Journals}
%

\maketitle

\begin{abstract}
Real-time semantic segmentation, which can be visually understood as the pixel-level classification task on the input image, currently has broad application prospects, especially in the fast-developing fields of autonomous driving and drone navigation. However, the huge burden of calculation together with redundant parameters are still the obstacles to its technological development. In this paper, we propose a Fast Bilateral Symmetrical Network (FBSNet) to alleviate the above challenges. Specifically, FBSNet employs a symmetrical encoder-decoder structure with two branches, semantic information branch and spatial detail branch. The Semantic Information Branch (SIB) is the main branch with semantic architecture to acquire the contextual information of the input image and meanwhile acquire sufficient receptive field. While the Spatial Detail Branch (SDB) is a shallow and simple network used to establish local dependencies of each pixel for preserving details, which is essential for restoring the original resolution during the decoding phase. Meanwhile, a Feature Aggregation Module (FAM) is designed to effectively combine the output of these two branches. Experimental results of Cityscapes and CamVid show that the proposed FBSNet can strike a good balance between accuracy and efficiency. Specifically, it obtains 70.9\% and 68.9\% mIoU along with the inference speed of 90 fps and 120 fps on these two test datasets, respectively, with only 0.62 million parameters on a single RTX 2080Ti GPU. The code is available at \url{https://github.com/IVIPLab/FBSNet}.
\end{abstract}


%
\IEEEpeerreviewmaketitle

\section{Introduction}
\label{sec1}

\IEEEPARstart{S}{emantic} segmentation, as one of the three fundamental computer vision tasks, is responsible for assigning a label to each pixel in an input image~\cite{zhang2019decoupled,gao2021mscfnet}. It can be viewed as a dense prediction task and what cannot be overlooked is the parameter burden. However, to embed it in real-world terminal equipment like augmented reality devices or autonomous driving chips, it is necessary to ensure that the model size and calculation cost of the proposed model are as small as possible. In comparison with developing large networks (e.g., VGGNet~\cite{simonyan2014very} and ResNet~\cite{he2016deep}) that pursue precision, designing real-time semantic segmentation structures is the proper choice that meets the needs of current real-world edge applications requiring fast interaction speed. 

\begin{figure}[t]
	\centerline{\includegraphics[width=9.5cm, trim=0 10 20 40]{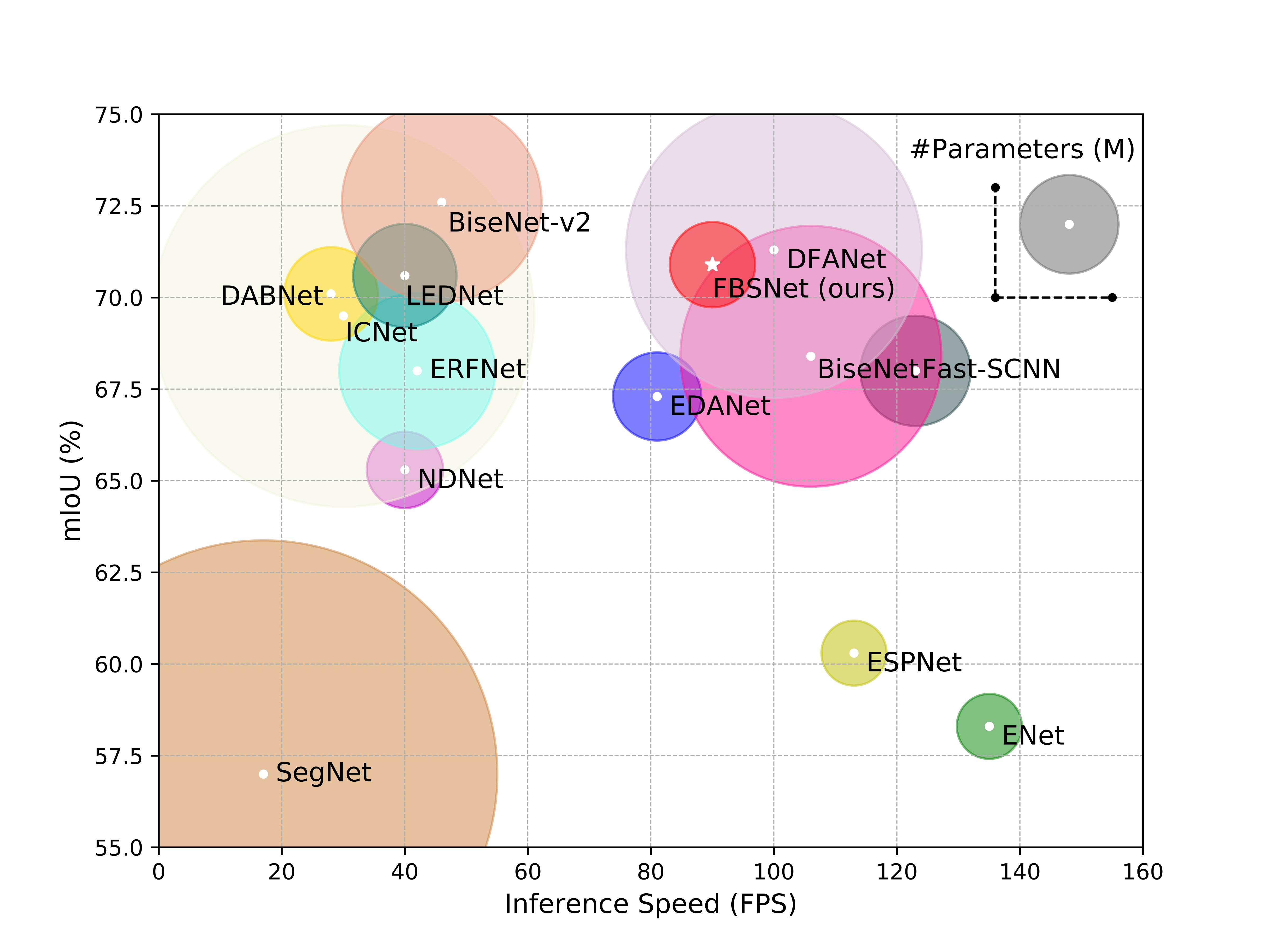}}
	\caption{Accuracy-Speed-Parameters comparisons on the Cityscapes. Our FBSNet achieves a good balance between the accuracy, model size, and inference speed of the model.}
	\label{Figure 1}
\end{figure}

\begin{figure*}[htbp]
	\centerline{\includegraphics[width=17.5cm]{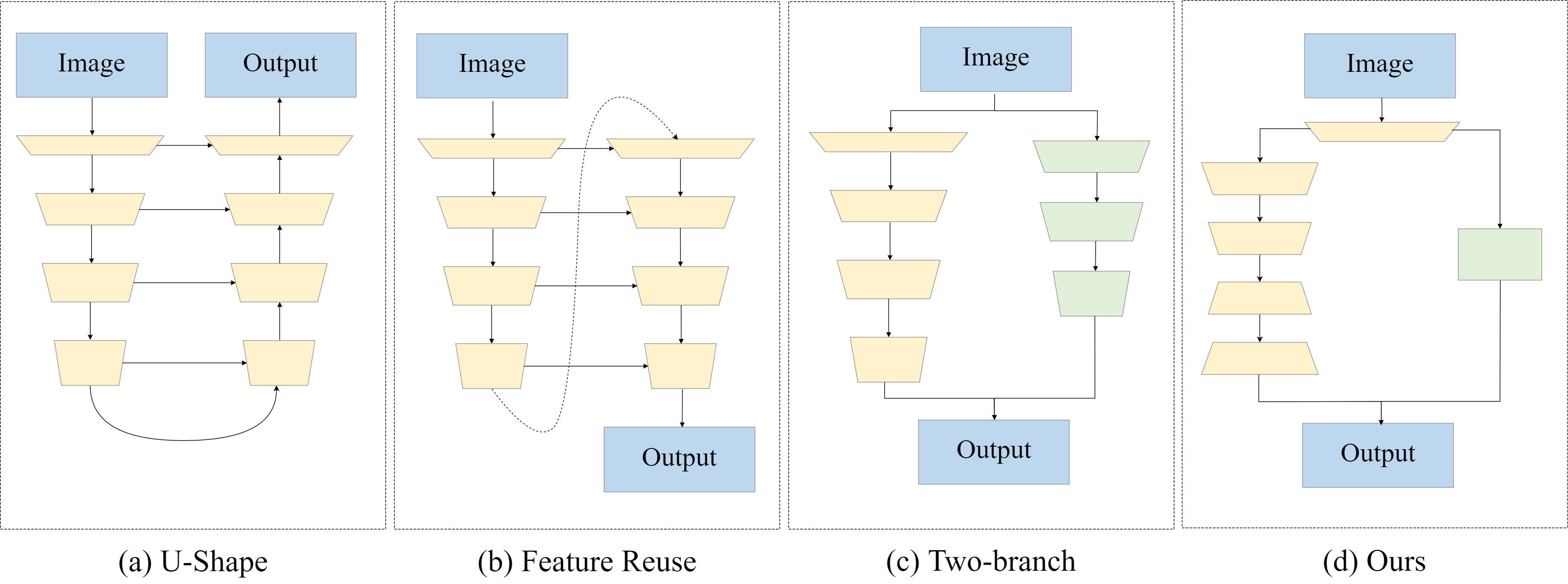}}
	\caption{Backbone comparison of commonly used two-branch image semantic segmentation models. Our FBSNet is significantly different from other models. Specifically, the two branches of FBSNet have different structures due to different tasks.}
	\label{Figure 2}
\end{figure*}

Recently, many lightweight real-time segmentation networks have been proposed to address the balance problem between accuracy and inference speed (Fig.~\ref{Figure 1}). For example, Paszke et al.~\cite{2016ENet} proposed an Efficient Neural Network (ENet), which drops the last stage to achieve a compact encoder-decoder framework. However, the disadvantage of this model is that the receptive field is too small to capture large objects. In order to collect multi-scale contextual information, Mehta et al.~\cite{mehta2018espnet} proposed the Efficient Spatial Pyramid Network (ESPNet), which adopts the efficient spatial pyramid module and convolution factorization strategy. ESPNet achieved better accuracy than ENet with similar parameters. The Image Cascade Network (ICNet~\cite{zhao2018icnet}) is improved from the Pyramid Scene Parsing Network (PSPNet~\cite{zhao2017pyramid}), which uses three cascading branches to process images. Nevertheless, this network is not suitable for low-resolution input images. The Bilateral Segmentation Network (BiSeNet~\cite{yu2018bisenet}) is a dual-path model, which contains two branches which can preserve the spatial details and obtain semantic information, respectively. Meanwhile, a feature fusion module was proposed to effectively aggregate the different features of these two branches. Based on BiSeNet, the fusion method of different features in BiSeNet-v2~\cite{2020BiSeNet} is further optimized. Although the effect of BiSeNet-v2 is much better than BiSeNet, the number of parameters and calculations has also increased a lot, which slows down the inference speed. DFANet~\cite{li2019dfanet} proposed a feature reuse strategy, which stacks Xception~\cite{chollet2017xception} models three times to yield the purpose of expanding the receptive field and the interactive integration of features. Fast-SCNN ~\cite{2019Fast} proposed a “learning to downsample” module and merged the two-branch setup in the encoder phase.

As we all know, when the depth of the network increases, the transmission of information will become quite difficult. At the same time, since the image resolution is down-sampled to a very low level, this will cause a lot of boundary information to be lost and cannot be recovered. Therefore, how to prevent information loss is very important. Meanwhile, how to make full use of the output features from the two-branch structure is also important. Although the aforementioned methods have achieved outstanding results, there is still the possibility of improvement in terms of accuracy and speed. Based on the above observations and considerations, we aim to explore a lightweight real-time semantic segmentation model. To this end, we propose a novel real-time network called Fast Bilateral Symmetrical Network (FBSNet). FBSNet is designed for effective inference, including higher accuracy and faster speed. FBSNet adopts a bilateral symmetrical encoder-decoder structure, including a Semantic Information Branch (SIB) and a Spatial Detail Branch (SDB). In order to reduce the parameters of the model, we do not use pre-trained models like ResNet~\cite{he2016deep} or VGG~\cite{simonyan2014very} as the backbone of the semantic information branch, but use our specially designed lightweight Bottleneck Residual Unit (BRU) to build this branch. BRU employs dilated factorized depth-wise separable convolutional layers to deepen the depth of the network and guarantees large receptive fields for extracting features, thus can drastically distill significant semantic information. Meanwhile, the SDB makes an effort to preserve spatial details at a small computational cost with a Detail Residual Module (DRM). Moreover, at the different stages of the SIB, we use channel attention modules to enhance the long-distance dependencies between channels. Correspondingly, to make up for the lost detailed information in SIB, we use the spatial attention module to generate an attention map for paying attention to the useful spatial information and ignoring useless information like noise in SDB. At the end of these two branches, we apply a Feature Aggregation Module (FAM) to combine and enhance the features at both semantic and spatial levels. As shown in Fig.~\ref{Figure 1}, our FBSNet achieves a good balance between the accuracy, model size, and inference speed of the model.

In summary, the contributions of this paper are as follows: 

\begin{itemize}
\item A lightweight Bottleneck Residual Unit (BRU) is proposed to drastically distill significant semantic information. BRU contains a small number of parameters and only needs less calculation cost, but can extract rich features with different receptive fields.

\item A Detail Residual Module (DRM) is proposed to better obtain shallow spatial features to compensate for the lost details in the semantic information branch. Meanwhile, a Feature Aggregation Module (FAM) is proposed to effectively fuse image features from different branches, with the goal to increase the global and local dependencies.

\item A novel Fast Bilateral Symmetrical Network (FBSNet) is presented for real-time image semantic segmentation. FBSNet is a symmetrical encoder-decoder structure, composed of a Semantic Information Branch (SIB) and a Spatial Detail Branch (SDB), which can effectively extract deep semantic information and preserve shallow boundary details, respectively. 
\end{itemize}

\begin{figure*}[ht]
	\centerline{\includegraphics[width=18.3cm]{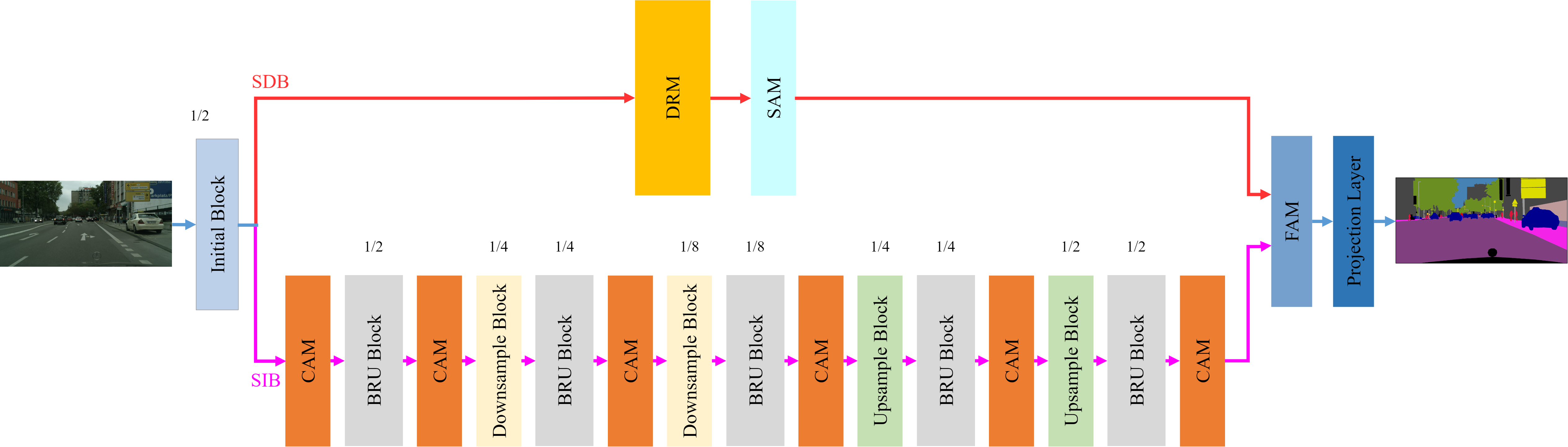}}
	\caption{The complete architecture of our proposed Fast Bilateral Symmetrical Network (FBSNet).}
	\label{Figure 3}
\end{figure*}

\section{Related Works} 
\label{sec2}

\subsection{Contextual Information}
\label{sec21}

Contextual information is crucial in the field of deep learning-based image processing methods. It contains a large number of image features that can be used for subsequent tasks like classification, detection, segmentation, and other tasks to predict high-quality results~\cite{gao2020constructing,gao2021hierarchical,zhang2018fast,wu2020cgnet}. For example, ParseNet~\cite{2015ParseNet} utilized global average pooling to generate weight feature maps for enhancing contextual information. Methods like~\cite{yang2018denseaspp,2017Rethinking} employed dilated convolution with various dilation rates to capture diversified contextual information. PSPNet~\cite{zhao2017pyramid}, inspired from DeepLab-V3~\cite{2017Rethinking}, applied spatial pyramid pooling (SPP) to capture global contexts. DFN~\cite{yu2018learning} encoded the global context by adding the global pooling on the head of the U-shape structure. Based on the above observations, it can be found that increasing the receptive field and acquiring a larger range of global contextual information is what every method cares about, and it is also beneficial to the final segmentation results.

\subsection{Spatial Information}
\label{sec22}

Spatial information is often ignored in the segmentation task. One is that with the deepening of the network layer and the use of consecutive down-sampling modules, the loss of spatial information is considered normal and unavoidable. The other is that restoring spatial information is a difficult task, which often consumes a lot of computing resources, so many methods ignore it~\cite{romera2017erfnet,shelhamer2017fully,wang2019lednet}. In fact, the preservation of spatial details has a positive effect on the prediction of the final results. Common methods, such as UNet~\cite{ronneberger2015u}, directly concatenate the shallow details and the features with corresponding resolutions. This operation will bring a large number of parameters and calculations, which is not good for the inference speed. In order to supplement the spatial detail information, DABNet~\cite{2019DABNet} inserted features after different multiples of downsampling at different stages in the phase of the encoder. ICNet~\cite{zhao2018icnet} applied a cascading method to supplement the missing boundary information. Some models~\cite{yang2018denseaspp,2017Rethinking} adopted dense connections, but cannot fulfill the real-time requirement. These models are all effective methods, but the decoder lacks the guidance of semantic information, resulting in unsatisfactory performance.

\subsection{General Structure} 
\label{sec23}

In the past ten years, many researchers have made a lot of efforts in network structure. However, many models still use the network structure of image classification (e.g., VGGNet~\cite{simonyan2014very}, ResNet~\cite{he2016deep}) as the backbone of the segmentation model. Although this is the easiest way, two problems cannot be ignored: one is that this type of model has too many parameters, the other is that frequent use of downsampling operations will cause a large amount of feature loss, which is not conducive to feature discrimination. To solve these problems, some representative network structures came into being, and many subsequent networks are improved based on previous methods, such as U-shape structure~\cite{ronneberger2015u,badrinarayanan2017segnet}, feature reuse~\cite{2015ParseNet,li2019dfanet}, and two-branch structure~\cite{yu2018bisenet,2020BiSeNet,2019Fast}. In Fig.~\ref{Figure 2}, we provide the structural comparison of some commonly-used image semantic segmentation models. Among them, the U-shape structure used the symmetrical encoder-decoder, whose strategy is to merge the feature maps of the corresponding stages. However, this kind of network is bound to bring huge additional computation. Feature reuse can enhance the network learning capacity and expand receptive fields by reusing high-level features. However, refining the spatial details is what prevents it from going further. Two-branch structure performs separate extraction of semantic information and spatial information in the encoder phase, and finally use a feature fusion method before prediction. But this method still lacks the interaction between the two branches, so there is still a lot of room for improvement.

In contrast to previous methods, the differences of our method are two-fold: (a) as for the two-branch structure, our main branch (semantic information branch) has a symmetrical encoder-decoder structure, while the spatial detail branch well preserves the shallow boundary details without the downsampling operations; (b) as for the feature fusion, to promote the representation capability of the fused features, except for the simple element-wise operation, we successively aggregate features at both semantic and spatial levels with negligible parameters and computational overhead.

\begin{figure*}[t]
	\centerline{\includegraphics[width=18cm]{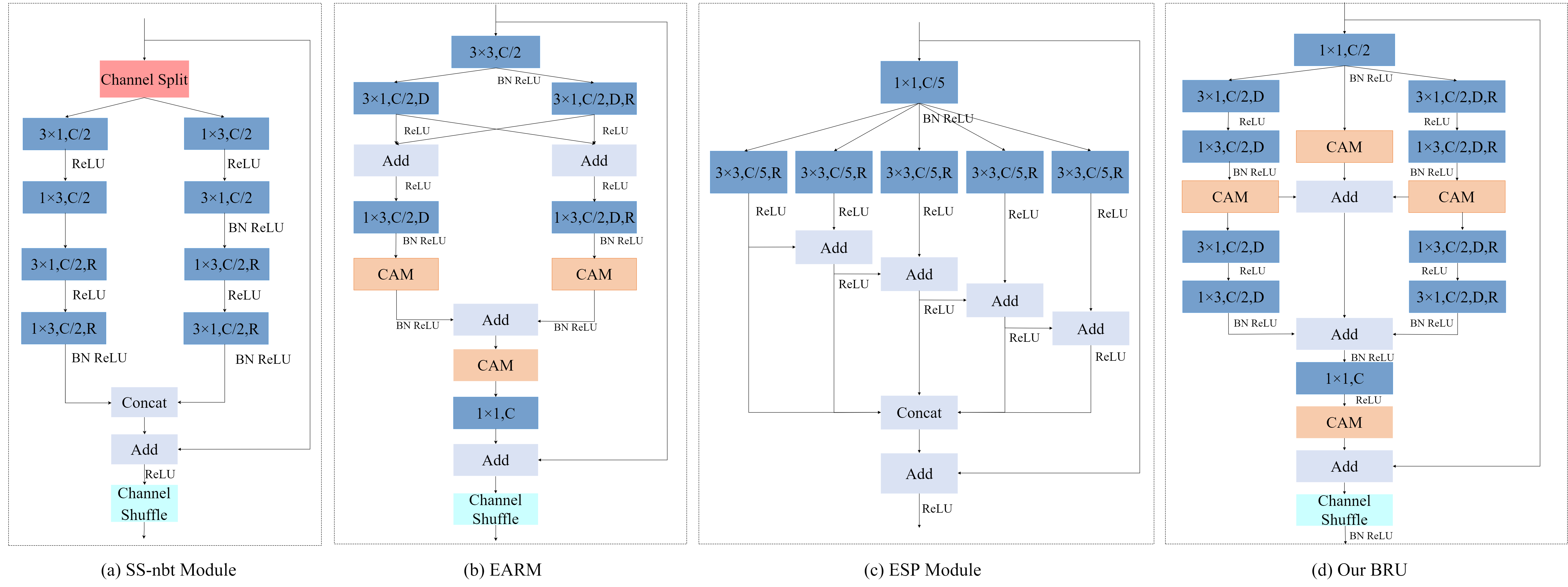}}
	\caption{Comparison of different residual modules. (a) SS-nbt module in LEDNet~\cite{wang2019lednet}, (b) EAR module in MSCFNet~\cite{gao2021mscfnet}, (c) Pyramid cascade module in ESPNet~\cite{mehta2018espnet}, and (d) is our proposed Bottleneck Residual Unit (BRU). Among them,  R, C, and D represent the dilation rate, the number of output channels, and the depth-wise separable convolution operation, respectively.}
	\label{Figure 4}
\end{figure*}

\section{Proposed Method}
\label{sec3}

In this part, we will introduce the proposed FBSNet with the following parts:
1)  Bottleneck Residual Unit (BRU);
2)	Semantic Information Branch (SIB) with channel attention module; 
3)	Spatial Detail Branch (SDB) with spatial attention module;
4)	Network with initial block and Feature Aggregation Module (FAM).

\subsection{Bottleneck Residual Unit (BRU)}
\label{sec31}

In Fig.~\ref{Figure 4}, we show some classic residual units that widely used in semantic segmentation. All of these modules are designed for feature feature extraction. Among them, (a) and (b) use the two-branch structure, one is responsible for local information extraction, while the other is dedicated to expanding the receptive field to obtain more contextual information. As for the pyramid cascade module in (c), it uses a series of dilation convolutions to gradually assemble multi-scale flexible representations. However, these models ignore the utilization of original features and the fusion of features from different branches is insufficient. To solve these problems, we proposed the Bottleneck Residual Unit (BRU). As shown in Fig.~\ref{Figure 4}, BRU is a three-branch module, the left branch is responsible for extracting local and short-distance feature information, the right branch is designed for enlarging the receptive field to acquire long-distance feature information, and the middle branch is dedicated to saving the input information. Specifically, we absorb the advantages of the bottleneck, which can not only ensure the model effect but also greatly reduce the calculation budget. Meanwhile, the convolution factorization strategy also be adopted, which means that a standard 2-dimensional convolution kernel $K \times K$ is factored into two 1-dimensional convolution kernels, i.e., $1 \times K$ and $K \times 1$ followed by the batch normalization~\cite{ioffe2015batch} and ReLU~\cite{nair2010rectified} operations. This strategy has been verified in many previous works such as Inception-v3~\cite{szegedy2016rethinking}, Xception~\cite{chollet2017xception}, MobileNet~\cite{sandler2018mobilenetv2,howard2019searching}, and ShuffleNet~\cite{zhang2018shufflenet,ma2018shufflenet}, which can significantly reduce the model parameters while maintaining the model performance. Meanwhile, the use of depth-wise separable convolution is based on the same reason and a $1 \times 1$ point-wise convolution is used to recover the final channel dependency. Dilation convolution is applied for enlarging the receptive fields. The rule is that as the network deepens, the dilation rate gradually becomes larger. At the end of the module, the information from three branches is merged together and a channel shuffle operation is applied to enhance the module representation capability. The complete operation of the module can be defined as follows:
\begin{equation}
{x_{out}} = {C_{1 \times 1}}\left( {{x_{in}}} \right),
\end{equation}
\begin{equation}
{y_{1,1}} = f_{CAM}\left( {{C_{1 \times 3}}\left( {{C_{3 \times 1}}\left( {{x_{out}}} \right)} \right)} \right),
\end{equation}
\begin{equation}
{y_1} = {C_{1 \times 3}}\left( {{C_{3 \times 1}}\left( {{y_{1,1}}} \right)} \right),
\end{equation}
\begin{equation}
{y_{2,1}} = f_{CAM}\left( {{C_{1 \times 3,r}}\left( {{C_{3 \times 1,r}}\left( {{x_{out}}} \right)} \right)} \right),
\end{equation}
\begin{equation}
{y_2} = {C_{3 \times 1,r}}\left( {{C_{1 \times 3,r}}\left( {{y_{2,1}}} \right)} \right),
\end{equation}
\begin{equation}
{y_3} = {y_{1,1}} + f_{CAM}(x_{out}) + {y_{2,1}},
\end{equation}
\begin{equation}
{y_{out}} = f_{shuffle}\left({f_{CAM}\left( {{C_{1 \times 1}}\left( {{y_1} + {y_2} + {y_3}} \right)} \right) + {x_{in}}} \right),
\end{equation}
where ${x_{in}}$ and ${y_{out}}$ mean the input and output of the BRU, respectively. ${y_1}$, ${y_2}$, and ${y_3}$ represent the output of the left, right, and middle branch, respectively. The first half of the outputs are represented by ${y_{1,1}}$ and ${y_{2,1}}$. Meanwhile, $f_{CAM}(\cdot)$ denotes the CAM operation, $C_{m \times n}$ denotes the convolutional layer with a $m \times n$ convolution kernel, and $f_{shuffle}(\cdot)$ is the channel shuffle operation. The activation function and batch normalization operations are omitted in the formulas.

\subsection{Semantic Information Branch (SIB)}
\label{sec32}

Recently, some modern approaches attempt to obtain sufficient receptive fields and enough contextual information by applying atrous spatial pyramid pooling, large kernel, or dense connection~\cite{zhao2017pyramid,yang2018denseaspp,2017Rethinking}. Although complex networks can bring good accuracy, the complex network and heavy computational overhead will make them incompatible with real-time requirements in real-world applications. The dual-branch structure network proposed by BiseNet~\cite{yu2018bisenet} used ResNet~\cite{he2016deep} as the backbone of the model to extract contextual information. The disadvantage is that it will take up a lot of GPU memory consumption and the used ResNet~\cite{he2016deep} is pre-trained by the ImageNet~\cite{krizhevsky2012imagenet} which cannot be regarded as a end-to-end model. 
Different from previous works, we build a deep Semantic Information Branch (SIB) with our proposed BRU. This not only ensures that more semantic information can be captured and larger receptive fields can be obtained, but also maintain the number of parameters and calculations is very low. Different stages of the convolutional neural networks have different representation abilities: the shallow stages preserve abundant spatial information, such as edges and corners but poor contextual information; while the deep stages have enough semantic consistency but a coarse prediction. Therefore, we set different dilation rates in BRU at different stages of the branch.

\textbf{Channel Attention Module (CAM).} Since the channels contain abundant feature information and interference noise, we use the Channel Attention Module (CAM) in both BRU and the semantic information branch to emphasize the features that need to be highlighted. Meanwhile, this method can suppress the interference noise thus facilitate feature extraction. Although CBAM~\cite{woo2018cbam}, Non-local~\cite{wang2018non}, DANet~\cite{fu2019dual}, and CCNet~\cite{huang2019ccnet} can bring good results, they are potentially not suitable for real-time segmentation tasks due to the complex calculation and a large number of parameters. Therefore, we use the lightweight attention method described in~\cite{wang2020eca}. CAM uses global average pooling to obtain global contexts and generates an attention map to guide the feature extraction with negligible computational cost, which is a good way to improve the model performance. This method has been generally explored in the tasks of detection, recognition, and segmentation. The process can be given as
\begin{equation}
M_c\left(X\right)=\sigma\left(C_{k\times k}\left(f_{Trans}\left( {f_{AvgPool}(X)}\right)\right)\right), \ \\
\end{equation}
where $M_c\in\mathbb{R}^{C\times 1\times 1}$ is the channel attention map, $X\in\mathbb{R}^{C\times H\times W}$ are the input features, $C_{k\times k}$ represents the standard convolution operation with kernel size $k$, $f_{AvgPool}(\cdot)$ represents the average pooling operation, $f_{Trans}(\cdot)$ denotes the compression and re-weighting operations, and $\sigma$ is the Sigmoid function.

\begin{figure}[t]
	\centerline{\includegraphics[width=9cm,trim=50 0 0 0]{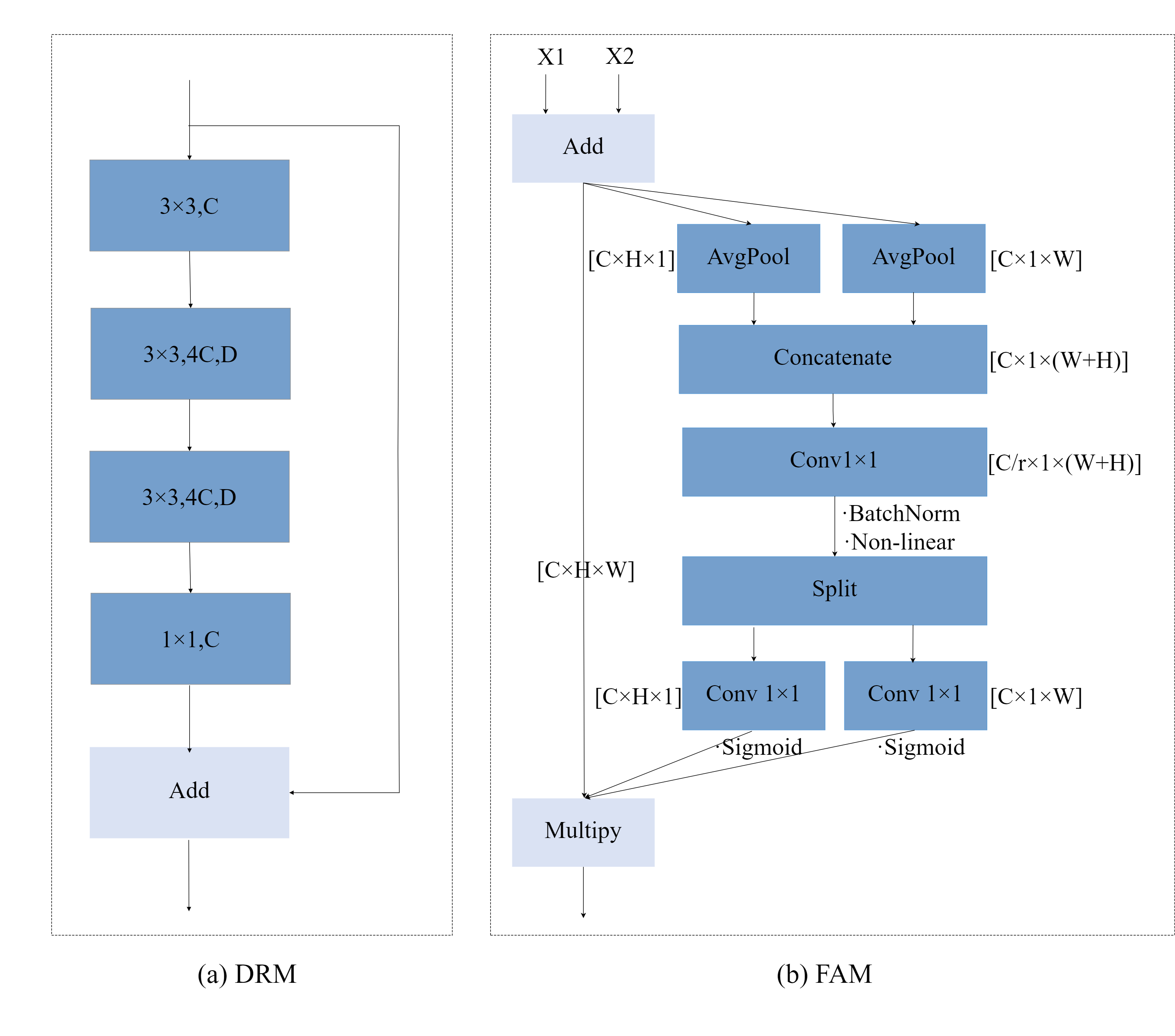}}
	\caption{The complete structure of (a) Detail Residual Module (DRM) and (b) Feature Aggregation Module (FAM).}
	\label{Figure 5}
\end{figure}

\textbf{Downsampling Block.} Inspired by the initial block in ENet~\cite{2016ENet}, we adopt it as our downsampling block. Through the downsampling operation, we can achieve larger receptive fields. However, too many downsampling operations will make the boundary details loss too serious to be restored. This is not conducive to feature extraction and utilization. Therefore, we take a compromised strategy and only use the downsampling module twice. That is, each time it is downsampled to 1/2 of the input features using a downsampling block. So we end up downsampling the image to 1/8 of the original given size.

\subsection{Spatial Detail Branch (SDB)}
\label{sec33}

Spatial information will be inevitably lost during the processing of the semantic information branch. The reason is that the extraction of deep-level semantic information and the preservation of shallow-level boundary information are a pair of contradictory relationships. To solve this problem, we design the spatial detail branch, which is actually a supplement to the detailed information lost by the semantic information branch to help the model achieve better accuracy in prediction process. Different from SIB, we only use a simple and effective Detail Residual Module (DRM, Fig.~\ref{Figure 5} (a)) and a Spatial Attention Module (SAM) in this branch. Among them, DRM is specially designed to supplement the details lost in the semantic branch, which consists of three $3 \times 3$ convolutional layers and one $1 \times 1$ point-wise convolutional layer. In order to obtain more features, we increase the number of channels of the second and third convolutional layers to 4 times (4C) the original input. Finally, we use a $1 \times 1$ convolutional layer to reduce the number of channels to C again. This operation can remove redundant features and extract effective features. In addition, to reduce the number of parameters and computational cost, we replace the later two $3 \times 3$ convolutional layers with the depth-wise separable convolution. Therefore, our DRM can extract rich shallow spatial features with few parameters and computational cost.

\begin{table*}[t]
\caption{Detailed architectural configuration of Fast Bilateral Symmetrical Network (FBSNet).}
\begin{center}	
\small
\setlength{\tabcolsep}{7mm}
\begin{tabular}{|c|c|cccc|}
\hline
Stage                     & Layer & \multicolumn{4}{c|}{Type (Size: $C \times H \times W$)}                                 \\ \hline\hline
\multirow{10}{*}{Encoder} & 1     & \multicolumn{2}{r}{Initial Block}  & \multicolumn{2}{l|}{($16 \times 256 \times 512$)} \\ \cline{2-6} 
                          & 2     & CAM                                & \multicolumn{1}{c|}{($16 \times 256 \times 512$)} & \multirow{8}{*}{DRM} & \multirow{8}{*}{\begin{tabular}[c]{@{}c@{}}($16 \times 256 \times 512$)\\($64 \times 256 \times 512$)\\ ($64 \times 256 \times 512$)\\ ($16 \times 256 \times 512$)\end{tabular}} \\ \cline{2-4}
                          & 3-5   & BRU (d=1) $\times 3$                & \multicolumn{1}{c|}{($16 \times 256 \times 512$)} &                      &                                                                                                            \\ \cline{2-4}
                          & 6     & CAM                                & \multicolumn{1}{c|}{($16 \times 256 \times 512$)}             &                      &                                                                                                            \\ \cline{2-4}
                          & 7     & Downsample Block                   & \multicolumn{1}{c|}{($64 \times 128 \times 256$)}             &                      &                                                                                                            \\ \cline{2-4}
                          & 8-10  & BRU (d=1) $\times 3$                & \multicolumn{1}{c|}{($64 \times 128 \times 256$)}             &                      &                                                                                                            \\ \cline{2-4}
                          & 11    & CAM                                & \multicolumn{1}{c|}{($64 \times 128 \times 256$)}             &                      &                                                                                                            \\ \cline{2-4}
                          & 12    & Downsample Block                   & \multicolumn{1}{c|}{($128 \times 64 \times 128$)}             &                      &                                                                                                            \\ \cline{2-4}
                          & 13-32 & \begin{tabular}[c]{@{}c@{}}{[}BRU (d=1)\\  BRU (d=2)\\  BRU (d=5)\\  BRU (d=9)\\  BRU (d=17){]} $\times 4$\end{tabular} & \multicolumn{1}{c|}{($128 \times 64 \times 128$)}      &                      &                                                                                                            \\ \cline{2-6} 
                          & 33    & CAM                                & \multicolumn{1}{c|}{($128 \times 64 \times 128$)}             &SAM                   & \multicolumn{1}{l|}{($16 \times 256 \times 512$)}                                                                                      \\ \hline
\multirow{8}{*}{Decoder}  & 34    & Upsample Block                     & \multicolumn{1}{c|}{($64 \times 128 \times 256$)}             & \multicolumn{2}{c|}{\multirow{6}{*}{}}                                                                                            \\ \cline{2-4}
                          & 35-37 & BRU (d=1) $\times 3$                & \multicolumn{1}{c|}{($64 \times 128 \times 256$)}             & \multicolumn{2}{c|}{}                                                                                                             \\ \cline{2-4}
                          & 38    & CAM                                & \multicolumn{1}{c|}{($64 \times 128 \times 256$)}             & \multicolumn{2}{c|}{}                                                                                                             \\ \cline{2-4}
                          & 39    & Upsample Block                     & \multicolumn{1}{c|}{($16 \times 256 \times 512$)}             & \multicolumn{2}{c|}{}                                                                                                             \\ \cline{2-4}
                          & 40-42 & BRU (d=1) $\times 3$                & \multicolumn{1}{c|}{($16 \times 256 \times 512$)}             & \multicolumn{2}{c|}{}                                                                                                             \\ \cline{2-4}
                          & 43    & CAM                                & \multicolumn{1}{c|}{($16 \times 256 \times 512$)}             & \multicolumn{2}{c|}{}                                                                                                     \\ \cline{2-6} 
                          & 44    & \multicolumn{2}{r}{FAM}           			& \multicolumn{2}{l|}{($16 \times 256 \times 512$)}                   \\ \cline{2-6} 
                          & 45    & \multicolumn{2}{r}{Projection Layer}  		& \multicolumn{2}{l|}{($19 \times 512 \times 1024$)}                  \\ \hline

\end{tabular}
\label{Table 1}
\end{center}
\end{table*}

\textbf{Spatial Attention Module (SAM).} Different from channel attention focusing on “what”, spatial attention pays more attention to “where”. Therefore, SAM is used to extract and preserve the shallow spatial information of the entire model. Specifically, the practice of spatial attention is to apply both max-pooling and average-pooling along the channel axis, then concatenate and convolve them by a standard convolution to generate an effective feature description. The process of spatial attention~\cite{woo2018cbam} can be described as follows:
\begin{equation}
M_s\left(X\right)=\sigma\left(C_{k\times k}\left(\left[ {f_{AvgPool}(X),f_{MaxPool}(X)} \right]\right)\right), \ \\
\end{equation}
where $M_s\in\mathbb{R}^{1\times H\times W}$ is the desired spatial attention map, $X\in\mathbb{R}^{C\times H\times W}$ are the input features, $C_{k\times k}$ represents the standard convolutional layer with kernel size $k$, $\left[ \right]$ means concatenation operation, $f_{AvgPool}(\cdot)$ represents the average pooling operation, $f_{MaxPool}(\cdot)$ represents the max pooling operation, and $\sigma$ is the Sigmoid function.

\subsection{Fast Bilateral Symmetrical Network (FBSNet)}
\label{sec34}

The whole network structure can be divided into three parts: initial block, dual-branch backbone, and feature aggregation module (the symmetry is mainly reflected in the resolution of the encoding and decoding procedure in SIB). The complete network structure can be found in Fig.~\ref{Figure 3} and the detailed architectural configuration can be found in TABLE~\ref{Table 1}. 

\textbf{Initial Block.} The initial block includes three $3 \times 3$ convolutional layers, where the stride of the first convolution is set to $2$ to collect initial features. Different from Fast-SCNN~\cite{2019Fast}, our initial block performs downsampling operation only once, thus the spatial information will be well preserved. In addition, in typical two-branches networks~\cite{yu2018bisenet,2020BiSeNet}, they often branched from the original input, where the feature information of the two branches is completely independent. In this paper, the initial block is used as the dividing point of two branches so that the semantic and spatial information are partially correlated, which facilitates subsequent feature merging.

\begin{table*}[!t]
\small
\setlength{\tabcolsep}{3mm}
\caption{Ablation studies on CamVid. SIB: Semantic Information Branch, SDB: Spatial Detail Branch, CAM: Channel Attention Module, FAM: Feature Aggregation Module, SAM: Spatial Attention Module.}
\begin{center}
\begin{tabular}{|ccccc|ccc|c|c|c|}
\toprule[1pt]
\multicolumn{1}{|c|}{\multirow{2}{*}{Models}} & \multicolumn{1}{c|}{\multirow{2}{*}{SIB}} & \multicolumn{1}{c|}{\multirow{2}{*}{SDB}} & \multicolumn{1}{c|}{\multirow{2}{*}{CAM}} & \multicolumn{1}{c|}{\multirow{2}{*}{SAM}} & \multicolumn{3}{c|}{Fusion}                                       & \multirow{2}{*}{Parameters (M)$\downarrow$} & \multirow{2}{*}{FLOPs (G)$\downarrow$} & \multirow{2}{*}{mIoU (\%)$\uparrow$} \\ \cline{6-8}
\multicolumn{1}{|c|}{}                        & \multicolumn{1}{c|}{}                     & \multicolumn{1}{c|}{}                     & \multicolumn{1}{c|}{}                     & \multicolumn{1}{c|}{}                     & \multicolumn{1}{c|}{Add} & \multicolumn{1}{c|}{Concatenate} & FAM &                                &                           &                           \\ \hline\hline
\multicolumn{11}{|l|}{(a) CAM}                                                                                                                        \\ \hline
FBSNet  		&\Checkmark    &    		&            &     			&      		&     		&     			&0.592270    &6.7433    &66.13          \\
FBSNet          &\Checkmark    &    		&\Checkmark  &     			&      		&     		&     			&0.592285    &6.7434    &66.72          \\ \hline \hline
\multicolumn{11}{|l|}{(b) FAM}                                                                                                                       \\ \hline
FBSNet      	&\Checkmark    &\Checkmark 	&\Checkmark  &  			&\Checkmark &     		&     			&0.614765    &9.6889    &67.87          \\
FBSNet          &\Checkmark    &\Checkmark 	&\Checkmark  &  			&      		&\Checkmark &     			&0.616093    &9.7745    &68.26          \\
FBSNet          &\Checkmark    &\Checkmark 	&\Checkmark  &  			&      		&     		&\Checkmark  	&0.615321    &9.6892    &68.58          \\ \hline \hline
\multicolumn{11}{|l|}{(c) SAM}                                                                                                                          \\ \hline
FBSNet          &\Checkmark    &\Checkmark  &\Checkmark  &     			&      		&     		&\Checkmark  	&0.615321    &9.6892    &68.58          \\
FBSNet          &\Checkmark    &\Checkmark  &\Checkmark  &\Checkmark  	&      		&     		&\Checkmark  	&0.615419    &9.7037    & \textbf{68.86}          \\ \bottomrule[1pt]
\end{tabular}
\label{Table 2}
\end{center}
\end{table*}

\begin{table*}[!t]
\small
\setlength{\tabcolsep}{4.2mm}
	\caption{Comparison with state-of-the-arts image semantic segmentation methods on the Cityscapes test dataset. Although our FBSNet did not achieve the best results on mIoU, the model achieves competitive results on the number of parameters, FLOPs, and FPS. In general, FSBNet achieves the best balance between model performance, model size, and inference time.}
	\begin{center}

		\begin{tabular}{|l|c|c|c|c|c|c|}
			\toprule[1pt]
			Methods                      							& Input Size  		   & Backbone   & Parameters (M) $\downarrow$ 	& FLOPs (G)$\downarrow$  & Speed (FPS)$\uparrow$  & mIoU (\%)$\uparrow$ \\ \hline\hline
			SegNet~\cite{badrinarayanan2017segnet}                  & $640 \times 360$     & VGG16      & 29.50                      	& 286.0      				& 17         			& 57.0     \\
			ENet~\cite{2016ENet}                   					& $512 \times 1024$    & No         & \textbf{0.36}              	& 3.8      				& 135        	& 58.3     \\
			SQNet~\cite{treml2016speeding}                  		& $1024 \times 2048$   & SqueezeNet & -             				& 270.0    				& 17         			& 59.8     \\
			ESPNet~\cite{mehta2018espnet}                 			& $512 \times 1024$    & ESPNet     & \textbf{0.36} 				& -        				& 113        			& 60.3     \\
			CGNet~\cite{wu2020cgnet}& $360 \times 640$   & No         & 0.50          				& 6.0     				& -        	 		& 64.8     \\
			NDNet~\cite{yang2020ndnet}                  			& $1024 \times 2048$   & No         & 0.50          				& 14.0     				& 40        	 		& 65.3     \\
			ESPNet-v2~\cite{mehta2019espnetv2}              		& $512 \times 1024$    & ESPNet-v2  & -             				& \textbf{2.7}      	& 80         			& 66.2     \\
			EDANet~\cite{lo2019efficient}                 			& $512 \times 1024$    & No         & 0.68         					& -        				& 81         			& 67.3     \\
			ADSCNet~\cite{wang2020adscnet} &$512 \times 1024$ & No & - & - & 77 & 67.5 \\
			ERFNet~\cite{romera2017erfnet}                 			& $512 \times 1024$    & No         & 2.10          				& -       			 	& 42         			& 68.0     \\
			Fast-SCNN~\cite{2019Fast}                 			& $1024 \times 2048$    & No         & 1.11          				& -       			 	& 123         			& 68.0     \\
			BiseNet~\cite{yu2018bisenet}                			& $768 \times 1536$    & Xception39 & 5.80          				& 14.8     				& 106        			& 68.4     \\
			BCPNet~\cite{hao2020bi} &$512 \times 1024$ &No & 0.61 & - & \textbf{250} & 68.4\\
			ICNet~\cite{zhao2018icnet}                  			& $1024 \times 2048$   & PSPNet50   & 26.50         				& 28.3     				& 30         			& 69.5     \\
			DABNet~\cite{2019DABNet}                 				& $1024 \times 2048$   & No         & 0.76          				& 10.5     				& 28         			& 70.1     \\
			LEDNet~\cite{wang2019lednet}                 			& $512 \times 1024$    & No         & 0.94          				& -        				& 40         			& 70.6     \\
			EdgeNet~\cite{han2020using} & $512 \times 1024$ & - & - & - & 31 & 71.0 \\
			DFANet~\cite{li2019dfanet}                 				& $1024 \times 1024$   & Xception   & 7.80          				& 3.4      				& 100        			& 71.3     \\
			FDDWNet~\cite{liu2020fddwnet} & $512 \times 1024$ & No & 0.80 & - & 60 & 71.5 \\
			MSCFNet~\cite{gao2021mscfnet} & $512 \times 1024$ & No  & 1.15 & 17.1 & 50  & 71.9 \\
			BiseNet-v2~\cite{2020BiSeNet}             				& $512 \times 1024$    & No         & 3.40         				& 21.2     				& 156         			& \textbf{72.6}     \\ \hline \hline
			FBSNet (ours)                 							& $512 \times 1024$    & No         & 0.62          				& 9.7      				& 90         			& 70.9     \\ \bottomrule[1pt]
		\end{tabular}
		\label{Table 3}
	\end{center}
\end{table*}

\textbf{Feature Aggregation Module (FAM).} How to effectively integrate the information of the semantic branch and the spatial branch is also the key issue of the two-branch structure. Directly element-wise add or concatenate them are the most widely used methods. However, these methods ignore the differences between the features provided by the two branches. To solve this problem, we used the strategy motivated by the latest attention mechanism~\cite{hou2021coordinate}. This method can not only capture cross-channel information but also grab information about direction and position perception. The most precious thing is that it is not computation-intensive, which means fewer parameters are in exchange for more gains, especially for dense predictions. Specifically, the global average pooling is factorized into a pair of 1D feature operations. Given an input ${X} \in {R^{C \times H \times W}}$, which is obtained by adding the two output features ${X_1} \in {R^{C \times H \times W}}$, ${X_2} \in {R^{C \times H \times W}}$ from the two branches. Along the horizontal and the vertical coordinates, we employ two spatial pooling kernels with size ($H$, 1) and (1, $W$) to encode the features, respectively. After that, we can get two transformations, which respectively merge features along the two spatial dimensions to correspondingly generate attention maps from two directions. One direction is responsible for capturing long-range dependencies, and the other helps to retain precise positional information. For the generated maps, we concatenate them and send them to a shared $1 \times 1$ convolutional layer for channel reduction yielding a $C/r \times 1 \times \left( {W + H} \right)$ output. The procedures can be formulated as follows:
\begin{equation}
F = {C_{1 \times 1}}\left( {\left[ {f_{AvgPool}^{H}\left( X \right),f_{AvgPool}^{W}\left( X \right)} \right]} \right), 
\end{equation}
where $[]$ represents the concatenation operation, $f_{AvgPool}^{H}(\cdot)$ and $f_{AvgPool}^{W}(\cdot)$ represent the pooling operation along the $H$ and $W$ directions. Batch normalization and non-linear activation function are also used before we split the intermediate features $F$ into two tensors, ${F^h} \in {R^{C/r \times H}}$ and ${F^w} \in {R^{C/r \times W}}$, respectively. After that, two $1 \times 1$ convolutional layers ${T_h}$ and ${T_w}$ are applied to restore ${F^h}$ and ${F^w}$ to have the same channel numbers as the input $X$. After passing through two Sigmoid functions, we expand the two outputs and take them as attention weights. The procedures can be formulated as follows: 
\begin{equation}
{K^h} = \delta \left( {{T_h}\left( {{F^h}} \right)} \right),
\end{equation}
\begin{equation}
{K^w} = \delta \left( {{T_w}\left( {{F^w}} \right)} \right).
\end{equation}

Finally, the output $Y$ of the FAM can be calculated as:
\begin{equation}
Y = \left( {{X_1} + {X_2}} \right) \times {K^h} \times {K^w}.
\end{equation}

Through this method, the features of the two branches can be fully integrated, and the feature information can be adaptively highlighted under the channel and spatial directions simultaneously. The complete structure of FAM is provided in Fig.~\ref{Figure 5} (b).

\begin{table*}[!t]
\caption{Per-class IoU (\%) results on the Cityscapes test set. “Avg” represents the average results of all these categories. Obviously, our FBSNet achieves the best mIoU results.}
\small
\setlength{\tabcolsep}{1.15mm}
\begin{tabular}{@{}|l|ccccccccccccccccccc|c|@{}}
\toprule[1pt]

Methods      & Tru  & Fen  & Wal  & Mot  & Tra  & Pol  & Rid  & TLi  & Ter  & Bic  & TSi  & Bus  & Ped  & Sid  & Bui  & Veg  & Car  & Sky  & Roa  & Avg  \\ \hline \hline
SegNet~\cite{badrinarayanan2017segnet}       & 38.1 & 29.0 & 28.4 & 35.8 & 44.1 & 35.7 & 42.8 & 39.8 & 63.8 & 51.9 & 45.1 & 43.1 & 62.8 & 73.2 & 84.0 & 87.0 & 89.3 & 91.8 & 96.4 & 57.0 \\ 
ENet~\cite{2016ENet}         & 36.9 & 33.2 & 32.2 & 38.8 & 48.1 & 43.4 & 38.4 & 34.1 & 61.4 & 55.4 & 44.0 & 50.5 & 65.5 & 74.2 & 75.0 & 88.6 & 90.6 & 90.6 & 96.3 & 58.3 \\ 
SQNet~\cite{treml2016speeding}        & 18.8 & 35.7 & 31.6 & 34.0 & 33.3 & 50.9 & 42.6 & 52.0 & 65.8 & 59.9 & 61.7 & 41.2 & 73.8 & 75.4 & 87.9 & 90.9 & 91.5 & 93.0 & 96.9 & 59.8 \\ 
ESPNet~\cite{mehta2018espnet}       & 38.1 & 36.1 & 35.0 & 41.8 & 50.1 & 45.0 & 40.9 & 35.6 & 63.2 & 57.2 & 46.3 & 52.5 & 67.0 & 77.5 & 76.2 & 90.8 & 92.3 & 92.6 & 97.0 & 60.3 \\ 
ESPNet-v2~\cite{mehta2019espnetv2}    & 53.0 & 42.1 & 43.5 & 44.2 & 53.2 & 49.3 & 53.1 & 52.6 & 66.8 & 59.9 & 60.0 & 65.9 & 72.9 & 78.6 & 88.8 & 90.5 & 91.8 & 93.3 & 97.3 & 66.2 \\ 
EDANet~\cite{lo2019efficient}       & 40.9 & 46.0 & 42.0 & 50.4 & \textbf{56.0} & 52.3 & 54.3 & 59.8 & 68.7 & 64.0 & 65.0 & 58.7 & 75.7 & 80.6 & 89.5 & 91.4 & 92.4 & 93.6 & 97.8 & 67.3 \\ 
ERFNet~\cite{romera2017erfnet}       & 50.8 & 48.0 & 42.5 & 47.3 & 51.8 & 56.3 & 57.1 & 59.8 & 68.2 & 61.7 & 65.3 & 60.1 & 76.8 & 81.0 & 89.8 & 91.4 & 92.8 & 94.2 & 97.7 & 68.0 \\ 
ICNet~\cite{zhao2018icnet}        & 51.3 & 48.9 & 43.2 & 53.6 & 51.3 & 61.5 & 56.1 & 60.4 & 68.3 & 70.5 & 63.4 & \textbf{72.7} & 74.6 & 79.2 & 89.7 & 91.5 & 92.6 & 93.5 & 97.1 & 69.5 \\ 
DABNet~\cite{2019DABNet}       & 52.8 & 50.1 & 45.5 & 51.3 & \textbf{56.0} & 59.3 & 57.8 & 63.5 & 70.1 & 66.8 & 67.7 & 63.7 & 78.1 & 82.0 & 90.6 & 91.8 & 93.7 & 92.8 & 97.9 & 70.1 \\ 
LEDNet~\cite{wang2019lednet}       & \textbf{64.4} & 49.9 & 47.7 & 44.4 & 52.7 & \textbf{62.8} & 53.7 & 61.3 & 61.2 & \textbf{71.6} & \textbf{72.8} & 64.0 & 76.2 & 79.5 & \textbf{91.6} & 92.6 & 90.9 & \textbf{94.9} & \textbf{98.1} & 70.6 \\ 
EdgeNet~\cite{han2020using}       & 50.0 & 50.6 & 45.4 & 55.3 & 52.5 & 62.6 & 61.1 & 67.2 & 69.7 & 67.7 & 71.4 & 60.9 & 80.4 & 83.1 &\textbf{91.6} & 92.4 &\textbf{94.3} & \textbf{94.9} & \textbf{98.1} & 70.6 \\
\hline
FBSNet
(ours) & 50.5 & \textbf{53.5} & \textbf{50.9} & \textbf{56.2} & 37.6 & 62.5 & \textbf{63.8} & \textbf{67.6} & \textbf{70.5} & 70.1 & 71.5 & 56.0 & \textbf{82.5} & \textbf{83.2} & 91.5 & \textbf{92.7} & 93.9 & 94.4 & 98.0 & \textbf{70.9} \\ \bottomrule[1pt]
\end{tabular}
\label{Table 4}
\end{table*}

\begin{figure*}[!t]
	\centerline{\includegraphics[width=18cm, height=7cm]{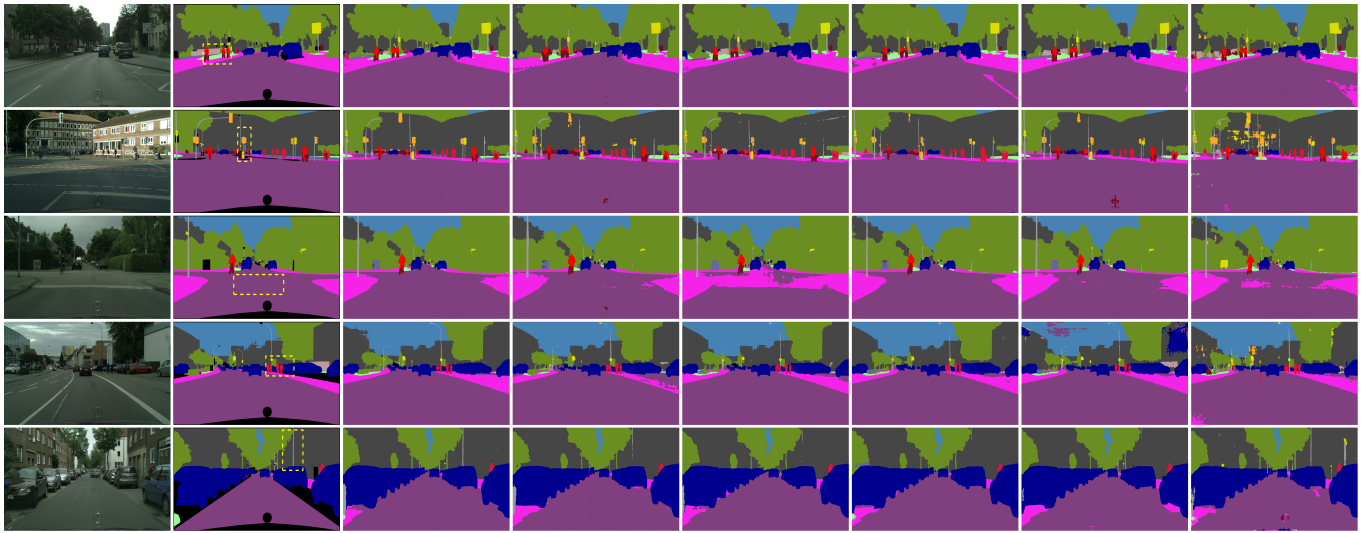}}
	\caption{Sample visual results on the Cityscapes validation set. From left to right: input samples, ground-truth references, segmentation outputs of the proposed FBSNet, LEDNet~\cite{wang2019lednet}, DABNet~\cite{2019DABNet}, ERFNet~\cite{romera2017erfnet}, NDNet~\cite{yang2020ndnet}, and ENet~\cite{2016ENet}. The region in the yellow dotted box can intuitively highlight the superiority of our method over others.}
	\label{Figure 6}
\end{figure*}

\section{Experiments}
\label{sec4}

In this section, we will introduce the benchmarks datasets we used, the parameter settings during training, the ablation experiments, as well as the comparison with other state-of-the-art image semantic segmentation methods to fully illustrate the advantages of our FBSNet.

\subsection{Datasets}
\label{sec41}

\textbf{Cityscapes.} The Cityscapes dataset records street scenes in 50 different European cities with the size of $2048 \times 1024$, containing 5,000 finely annotated images and 19,998 coarsely annotated images. The dataset involves 30 categories, and we only employ 19 of them with fine-annotated images for training and testing, which is divided into three parts: 2975 samples for training, 500 samples for validation, and 1525 samples for testing.

\textbf{CamVid.} The CamVid dataset is another vehicle applications dataset collected from video sequences. It has 11 categories and 701 finely annotated images that are split into 367 training samples, 101 validation samples, and 233 testing samples with the same size of $960 \times 720$.

\subsection{Implementation details}
\label{sec42}

We implement our model with the PyTorch framework and all our experiments are performed on a single RTX 2080 Ti GPU. For Cityscapes, we train our model from scratch by utilizing cross-entropy loss function with OHEM, as well as the stochastic gradient descent (SGD) method with momentum 0.9 and the related weight decay $1 \times 10^{-4}$. As for the CamVid dataset, due to different input resolutions, we adapt the optimization method to Adam with momentum 0.9 and related weight decay $2 \times 10^{-4}$. At the same time, to stabilize the training, the ‘poly’ policy is also utilized for the learning rate strategy. The initial learning rate is configured as $4.5 \times 10^{-2}$ and $1 \times 10^{-3}$ for Cityscapes and CamVid, respectively. The learning rate is generally related to the iteration and can be calculated as $lr_{initial}\times\left(1-\frac{iteration}{max\_iteration}\right)^{0.9}$. Both datasets are trained for 1000 epochs with a batch size of $4$ for Cityscapes and $6$ for CamVid.

\subsection{Ablation Studies}
\label{sec43}

The main purpose of the ablation studies conducted on the CamVid test set is to verify the performance improvement of each module in the entire network. Similar to the controlled variable methods in physics experiments, we add modules step by step to see their impact on the prediction results, the number of parameters, and the amount of computation. All results are provided in TABLE~\ref{Table 2}. 

\textbf{CAM.}
As we can observe from the first two lines of the data field in TABLE~\ref{Table 2}, the prediction results of the network are worse without using CAM. In other words, the CAM can provide a 0.6\% improvement in accuracy with almost no increase in the number of parameters or computation. This experiment fully proves the effectiveness of CAM.

\textbf{FAM.}
Feature fusion methods have always been the key research topic for multi-semantic aggregation. Among them, the ``Add'' and ``Concatenate'' operations are the most widely used methods. Therefore, we provide the comparison of ``Add'', ``Concatenate'', and FAM in TABLE~\ref{Table 2}. According to the table, we can clearly observe that our FAM achieves the best performance of 68.58\%, which is 0.71\% and 0.32\% better than that of the ``Add'' and ``Concatenate'' operations, respectively. Compared with the “Add” operation, FAM only increases negligible parameters (0.000556M) and FLOPs (0.0003G). This is entirely acceptable. Moreover, compared with the “Concatenate” operation, FAM achieves better results with fewer parameters and FLOPs. This is a big breakthrough that can effectively improve the performance of the model without increasing the complexity of the model. Comprehensive consideration of these indicators can prove the effectiveness of our proposed FAM.

\textbf{SAM.}
This part is to verify the benefits of SAM in enhancing spatial information. According to TABLE~\ref{Table 2}, we can see that SAM can take up only 0.01G of computation and very few parameters, resulting in a performance improvement from 68.58\% to 68.86\%.

As can be seen from the above three ablation experiments, each module plays a unique role in the network. It can also be found that due to the addition of SDB, the computation of the whole network is increased by nearly a half while the performance is improved from 66.72\% to 67.87\%. The reason why many methods are reluctant to use it is that their semantic branches are already so computationally heavy that there is no room left to consider the boundary detail information. On the contrary, due to the lightweight structure of our SIB, we can integrate the SIB and SDB to achieve promising performance.

\begin{table}[t]
\small
\setlength{\tabcolsep}{0.6mm}
	\caption{Comparison with state-of-the-arts image semantic segmentation methods on the CamVid test set.}
	
		\begin{tabular}{|l|c|c|c|c|}
			\toprule[1pt]
			Methods   									& Input Size 					& Backbone 			& Parameters (M)$\downarrow$ 	 & mIou (\%)$\uparrow$ \\ \hline\hline
			ENet~\cite{2016ENet}      					& $360 \times 480$    			& No       			& \textbf{0.36} 				 & 51.3     \\ 
			SegNet~\cite{badrinarayanan2017segnet}    	& $360 \times 480$    			& VGG16    			& 29.50         				 & 55.6     \\ 
			NDNet~\cite{yang2020ndnet}     				& $360 \times 480$    			& No       			& 0.50          				 & 57.2     \\
			DFANet~\cite{li2019dfanet}    				& $720 \times 960$   		    & Xception 			& 7.80         					 & 64.7     \\ 
			Dilation8~\cite{pohlen2017full} 			& $720 \times 960$    			& VGG16    			& 140.80        				 & 65.3     \\
			CGNet~\cite{wu2020cgnet}& $360 \times 480$              & No 				& 0.50							 & 65.6     \\
			BiseNet~\cite{yu2018bisenet}   				& $720 \times 960$    			& Xception39 		& 5.80         				 & 65.6     \\
			DABNet~\cite{2019DABNet} 					& $360 \times 480$              & No 				& 0.76							 & 66.4     \\
			FDDWNet~\cite{liu2020fddwnet}         & $360 \times 480$    			& No 			& 0.80        				 	 & 66.9     \\
			ICNet~\cite{zhao2018icnet}     				& $720 \times 960$    			& PSPNet50 			& 26.50        				 	 & 67.1     \\
			BCPNet~\cite{hao2020bi}   				& $720 \times 720$    			& No 			& 0.61         				 	 & 67.8     \\
			BiseNet-v2~\cite{2020BiSeNet}   				& $720 \times 960$    			& ResNet18 			& 49.00         				 	 & 68.7     \\
			MSCFNet~\cite{gao2021mscfnet} & $360 \times 480$    			& No       	& 1.15         	 & \textbf{69.3} \\
			\hline \hline
			FBSNet (ours)   									& $360 \times 480$    			& No       			& 0.62         	 & 68.9    \\ \bottomrule[1pt]
		\end{tabular}
		\label{Table 5}
	
\end{table}

\begin{figure}[t]
	\centerline{\includegraphics[width=8.8cm]{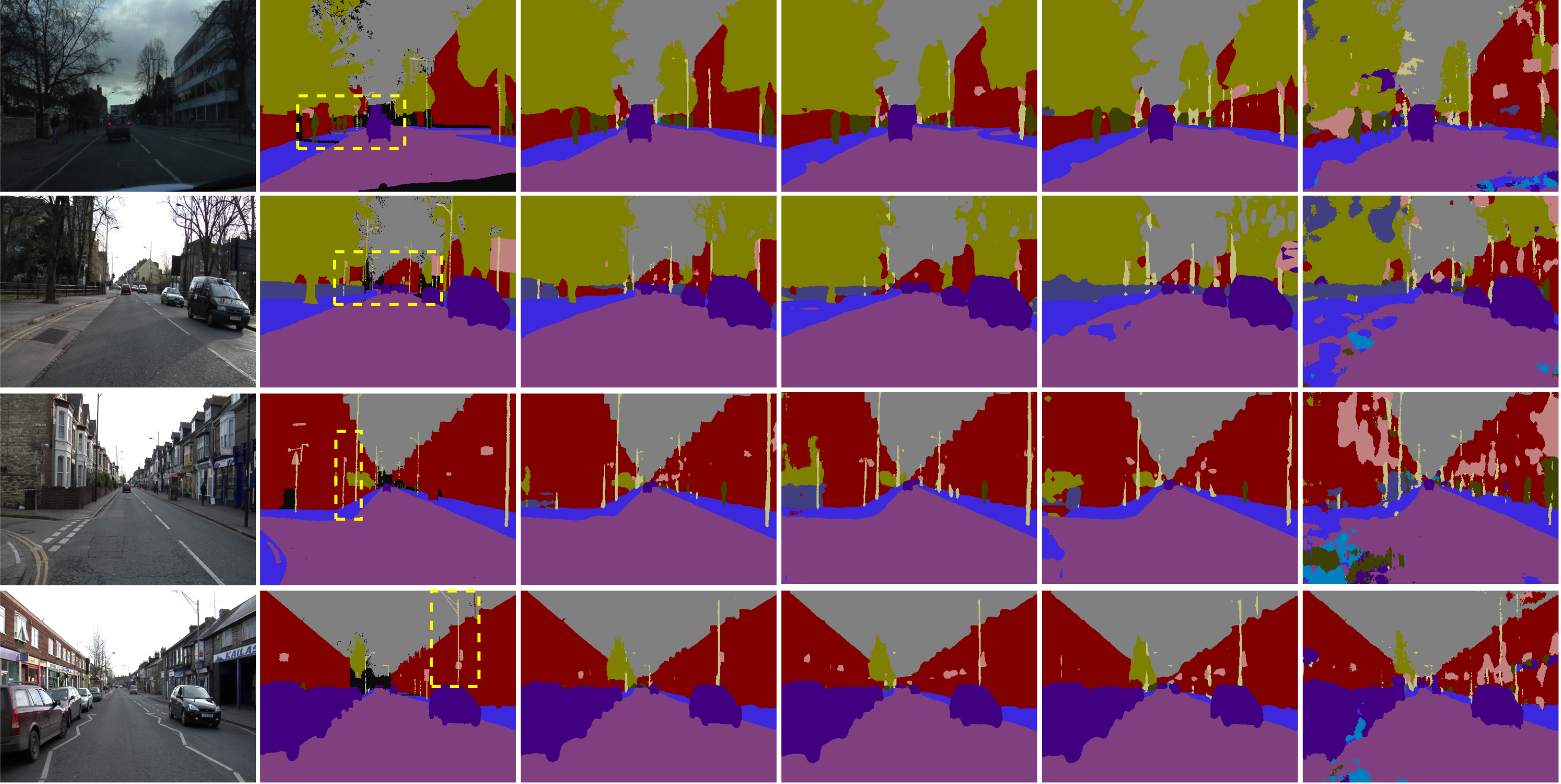}}
	\caption{Visual comparison on CamVid. From left to right: input samples, ground-truth references, segmentation outputs of the proposed FBSNet, DABNet~\cite{2019DABNet}, DFANet~\cite{li2019dfanet}, NDNet~\cite{yang2020ndnet}, and ENet~\cite{2016ENet}. The region in the yellow dotted box can intuitively highlight the superiority of our method over others.}
	\label{Figure 7}
\end{figure}

\subsection{Comparison with SOTA Methods}
\label{sec44}

During the training phase, we crop the size of original images and corresponding labels to $512 \times 1024$ for Cityscapes and $480 \times 360$ for CamVid as input to reduce the computational consumption and the GPU memory occupation. For testing, we use the original size of the datasets for inference. For a fair comparison, we do not employ any other post-process operations, like conditional random field (CRF). The evaluation indicators of the model are mainly from four aspects: the mean intersection over union (mIoU), the number of parameters, the number of float-point operations (FLOPs), and the inference speed (FPS).

\textbf{Results on Cityscapes:} In TABLE~\ref{Table 3}, we provide a quantitative comparison with other state-of-the-art image semantic segmentation methods on the Cityscapes test dataset. Moreover, we provide the per-class IoU (\%) results on the Cityscapes in TABLE~\ref{Table 4}. According to these results, we can observe that when encountered with fewer parameters, our network can well ensure better accuracy and faster inference speed. The models with the same number of parameters as our method do not have the same effects, and the models with the same effects will need more parameters than ours. Specifically, from the perspective of ``Parameters'', ENet~\cite{2016ENet}, ESPNet~\cite{mehta2018espnet}, and NDNet~\cite{yang2020ndnet} have fewer parameters but their segmentation accuracy is 5.6\% lower than our FBSNet, which is a big gap in the segmentation field. This is in the case that our number of parameters is only 0.26M more than theirs. From the perspective of ``FLOPs'', the computation cost of ESPNet-v2~\cite{mehta2019espnetv2} is 7G smaller than ours at the cost of 4.7\% segmentation performance penalty. From the perspective of ``Speed'', our method is in a position above the medium level at 90 FPS. While in terms of ``mIoU'', there are several methods have higher performance than our method, such as DFANet~\cite{li2019dfanet} and BiseNet-v2~\cite{2020BiSeNet}. Indeed, DFANet~\cite{li2019dfanet} is an outstanding model, which achieves slightly better mIoU result than our FBSNet. However, we should not ignore that the number of parameters of DFANet is 7.8M (M= Million), while the number of ours model is 0.62M, which is only 1/12 of DFANet. Such a large-scale DFANet is difficult to be promoted in practical applications. As for the inference speed, the results of the two models are very close (DFANet: 100 vs. FBSNet: 90). The reason why FBSNet has fewer parameters but slower inference speed is that the attention mechanisms are used in FBSNet. These attention mechanisms will bring some computational overhead, resulting in slower inference speed. Fortunately, this effect is acceptable. To better compare these models, we provide the Accuracy-Speed-Parameters comparisons in Fig.~\ref{Figure 1}. Obviously, our FBSNet achieves the best trade-off between the accuracy, model size, and inference speed of the model. Moreover, we provide the visual comparisons with LEDNet~\cite{wang2019lednet}, DABNet~\cite{2019DABNet}, ERFNet~\cite{romera2017erfnet}, NDNet~\cite{yang2020ndnet}, and ENet~\cite{2016ENet} in Fig.~\ref{Figure 6}. Through a horizontal comparison of the segmentation results, we can also qualitatively see the advantages of our proposed method. The yellow dotted boxes in the ground truth distinctly depict that our method is obviously superior to other methods.

\textbf{Results on CamVid:} 
To further verify the performance of FBSNet, we also provide a quantitative comparison with other methods on the CamVid test dataset in TABLE~\ref{Table 5}. 
According to the table, we can observe that (a) FBSNet achieves the second-best segmentation results, which is slightly worse than MSCFNet. However, it should be noticed that FBSNet requires only half the amount of parameters and FLOPs of MSCFNet, and its inference speed is twice that of MSCFNet; (b) Compared with methods of similar model size, the performance of FBSNet has been significantly improved; (c) Compared with other large models, FBSNet achieves better results with fewer parameters. Similarly, we provide the visual comparison results of these methods on the CamVid test dataset in Fig.~\ref{Figure 7}. Obviously, our FBSNet still achieves the best segmentation results. All the above experiments have fully proved the effectiveness and excellence of FBSNet, which strikes a good balance between accuracy and efficiency.

\section{Conclusions}
In this paper, we proposed a Fast Bilateral Symmetrical Network (FBSNet) for real-time image semantic segmentation. Compared to some top-accurate methods struggling for complex architecture and extreme lightweight networks aiming at compressing convolutional layers, our solution mainly focuses on getting a relatively good balance between the accuracy, model size, and inference speed with elaborately designed modules. Especially, our proposed Bottleneck Residual Unit (BRU) can enlarge the receptive field to distill significant semantic information comprehensively. Our extensive investigations have shown that spatial information is critical to recovering the resolution during the decoder process. Therefore, we designed a symmetrical encoder-decoder model, which is composed of a Semantic Information Branch (SIB) and a Spatial Detail Branch (SDB). These two branches can extract deep semantic information and preserve shallow boundary details, respectively. In summary, our model, which is only 0.62M, can process a $512 \times 1024$ input size and achieves 70.9\% at a speed of 90 FPS on Cityscapes test data with a single RTX 2080Ti GPU. Extensive experiments have revealed that our proposed FBSNet achieved a good balance between the accuracy and efficiency of the model.

\bibliographystyle{IEEEtran}
\bibliography{reference}

%

\begin{IEEEbiographynophoto}{Guangwei Gao}
 (IEEE Member) received the Ph.D. degree in pattern recognition and intelligence systems from the Nanjing University of Science and Technology, Nanjing, in 2014. He was a Project Researcher with the National Institute of Informatics, Japan, in 2019. He is currently an Associate Professor in Nanjing University of Posts and Telecommunications. His research interests include pattern recognition, and computer vision. He has published more than 40 scientific papers in IEEE TIP, IEEE TCSVT, IEEE TITS, IEEE TMM, PR, AAAI, etc, and served as a reviewer for journals and conferences including IEEE TPAMI, IEEE TMM, IEEE TCSVT, CVPR, ICCV, ECCV, AAAI, etc. Personal website: \textit{https://guangweigao.github.io}.
\end{IEEEbiographynophoto}

\begin{IEEEbiographynophoto}{Guoan Xu}
 received the B.S degrees in Measurement Control Technology and Instrumentation from Changshu Institute of Technology, Jiangsu, China, in 2019. He is currently pursuing the M.S. degree with the College of Automation \& College of Artificial Intelligence, Nanjing University of Posts and Telecommunications. His research interests include image semantic segmentation.
\end{IEEEbiographynophoto}

\begin{IEEEbiographynophoto}{Juncheng Li}
received the Ph.D. degree in Computer Science and Technology from East China Normal University, Shanghai, China, in 2021. He is currently a Postdoctoral Fellow at the Center for Mathematical Artificial Intelligence, The Chinese University of Hong Kong. His main research interests include image restoration, computer vision, deep learning, and medical image processing. He has published more than 20 scientific papers in IEEE TIP, IEEE TNNL, IEEE TMM, IEEE TCSVT, PR, ICCV, ECCV, AAAI, etc.
\end{IEEEbiographynophoto}

\begin{IEEEbiographynophoto}{Yi Yu}
(Member, IEEE) received the Ph.D. degree in information and computer science from Nara Women’s University, Japan. He is currently an Assistant Professor with the National Institute of Informatics (NII), Japan. Before joining NII, she was a Senior Research Fellow with the School of Computing, National University of Singapore. Her research covers large-scale multimedia data mining and pattern analysis, location-based mobile media service and social media analysis. She and her team received the best Paper Award from the IEEE ISM 2012, the 2nd prize in Yahoo Flickr Grand Challenge 2015, were in the top winners (out of 29 teams) from ACM SIGSPATIAL GIS Cup 2013, and the Best Paper Runner-Up in APWeb-WAIM 2017, recognized as finalist of the World’s FIRST 10K Best Paper Award in ICME 2017.
\end{IEEEbiographynophoto}

\begin{IEEEbiographynophoto}{Huimin Lu}
(Senior Member, IEEE) received the B.S. degree in electronics information science and technology from Yangzhou University in 2008, the M.S. degree in electrical engineering from the Kyushu Institute of Technology and Yangzhou University in 2011, and the Ph.D. degree in electrical engineering from the Kyushu Institute of Technology in 2014. From 2013 to 2016, he was a JSPS Research Fellow (DC2, PD, and FPD) with the Kyushu Institute of Technology. He is currently an Assistant Professor with the Kyushu Institute of Technology and an Excellent Young Researcher of MEXT-Japan. His research interests include computer vision, robotics, artificial intelligence, and ocean observing.
\end{IEEEbiographynophoto}

\begin{IEEEbiographynophoto}{Jian Yang}
(Member, IEEE) received the PhD degree from Nanjing University of Science and Technology (NUST), on the subject of pattern recognition and intelligence systems in 2002. In 2003, he was a postdoctoral researcher at the University of Zaragoza. From 2004 to 2006, he was a Postdoctoral Fellow at Biometrics Centre of Hong Kong Polytechnic University. From 2006 to 2007, he was a Postdoctoral Fellow at Department of Computer Science of New Jersey Institute of Technology. Now, he is a Chang-Jiang professor in the School of Computer Science and Engineering of NUST. He is the author of more than 100 scientific papers in pattern recognition and computer vision. His papers have been cited more than 4000 times in the Web of Science, and 9000 times in the Scholar Google. His research interests include pattern recognition, computer vision and machine learning. Currently, he is/was an Associate Editor of Pattern Recognition Letters, IEEE Trans. Neural Networks and Learning Systems, and Neurocomputing. He is a Fellow of IAPR.
\end{IEEEbiographynophoto}





\end{document}